\documentclass{article}
\PassOptionsToPackage{numbers,sort,compress}{natbib}
\usepackage[preprint]{neurips_2026}
\makeatletter
\renewcommand{\@noticestring}{Preprint. Work in progress.}
\makeatother
\usepackage{xcolor}
\definecolor{CitationBlue}{HTML}{1F5AA6}
\usepackage[colorlinks=true,citecolor=CitationBlue,linkcolor=black,urlcolor=CitationBlue]{hyperref}
\usepackage{url}
\usepackage{xurl}
\usepackage{booktabs}
\usepackage{multirow}
\usepackage{graphicx}
\usepackage{array}
\usepackage{placeins}
\usepackage{amsthm}

\usepackage{amsmath,amsfonts,bm}

\def\eqref#1{equation~\ref{#1}}

\def\1{\bm{1}}

\DeclareMathAlphabet{\mathsfit}{\encodingdefault}{\sfdefault}{m}{sl}
\SetMathAlphabet{\mathsfit}{bold}{\encodingdefault}{\sfdefault}{bx}{n}

\newcommand{\method}{\textsc{SPO++}}
\newcommand{\spobase}{\textsc{SPO}}
\newtheorem{remark}{Remark}

\title{SPO++: Stream-Aligned Policy Optimization for Asynchronous Agentic RL}
\newcommand{\PublicPDFAuthors}{Kai Ruan, Jinghao Lin, Qianshan Wei, Ziqi Zhou, and Zihe Huang}
\author{
  \textbf{Kai Ruan}$^{1,\dagger}$\hspace{0.12em}\thanks{Corresponding author: \href{mailto:kairuan@ruc.edu.cn}{kairuan@ruc.edu.cn}.} \quad
  \textbf{Jinghao Lin}$^{2,\dagger}$ \quad
  \textbf{Qianshan Wei}$^{3}$ \\
  \textbf{Ziqi Zhou}$^{4}$ \quad
  \textbf{Zihe Huang}$^{5}$ \\[2pt]
  \normalfont
  $^1$Gaoling School of Artificial Intelligence, Renmin University of China \\
  $^2$Independent Researcher \quad
  $^3$Institute of Automation, Chinese Academy of Sciences \\
  $^4$Duke University \quad
  $^5$Institute of Computing Technology, Chinese Academy of Sciences \\
  $^\dagger$Equal contribution
}

\hypersetup{pdftitle={SPO++: Stream-Aligned Policy Optimization for Asynchronous Agentic RL},pdfauthor={\PublicPDFAuthors}}
\newcommand{\PaperAbstract}{%
Group-relative reinforcement learning waits for sibling rollouts of the same prompt, which is costly for long and variable tool-use trajectories.  Single-stream Policy Optimization (SPO) removes this dependency with a persistent prompt-level value estimate, but its recipe whitens one advantage per trajectory before optimizing a token-mean actor loss.  We show that trajectory centering generally does not center the token-weighted quantity consumed by the actor, and fix the mismatch by standardizing terminal-outcome advantages under the action-token measure.  We additionally organize prompt evidence by the policy event that generated it rather than learner receipt order.  Across matched runs on ALFWorld at two model scales and on Math-TIR, \method{} improves online learning efficiency over SPO.  A paired ablation identifies action-token-measure normalization as the strongest tested component.}

\begin{document}
\maketitle
\begin{abstract}
  \PaperAbstract
\end{abstract}

\section{Introduction}
\label{sec:introduction}

Reinforcement learning with verifiable outcomes is increasingly used to train language-model reasoning and tool use.  GRPO estimates a critic-free advantage from multiple outputs sampled for one prompt, and later large-scale recipe refinements improve the stability of this family \citep{shao2024deepseekmath,yu2025dapo}.  For agents, however, the group is also a barrier: environment interaction, tool calls, and failed trajectories create long completion-time tails, yet the update waits for the slowest sibling.

SPO replaces the ephemeral group baseline with persistent prompt-level evidence, enabling one online rollout per prompt \citep{xu2025spo}.  This is attractive for asynchronous execution, but exposes two consistency questions.  First, a sequential prompt tracker observes learner receipt order rather than the policy event that generated an outcome.  Second, the original SPO recipe whitens equally weighted trajectory advantages and subsequently broadcasts each scalar over a variable number of response tokens.  If the actor loss is averaged over tokens, response length silently changes the center seen by optimization.

We study these two mismatches in \method{}.  The method retains SPO's terminal trajectory outcome and single-rollout dependency, but (i) freezes prompt values at dispatch and records returned evidence using a policy-event coordinate, and (ii) normalizes under the same action-token measure consumed by the loss.  Together, these changes align the persistent baseline with the policy clock and the scalar advantage with the actor-loss measure.

\begin{figure}[t]
  \centering
  \includegraphics[width=\linewidth]{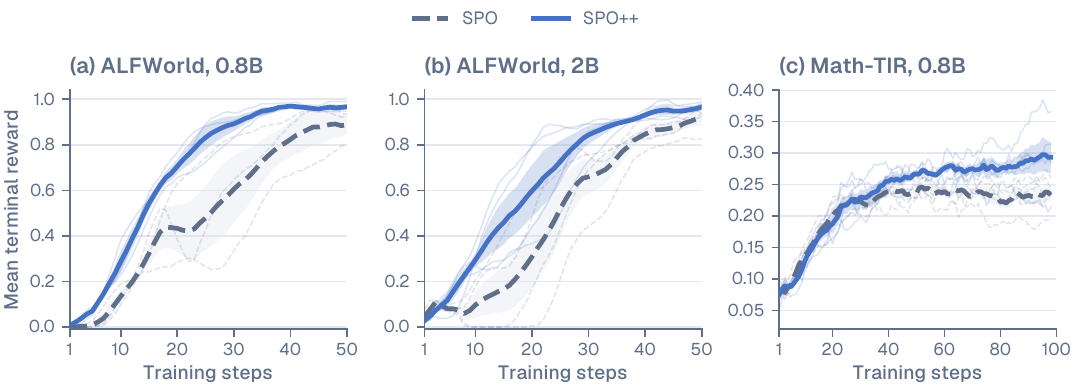}
  \caption{\method{} acquires reward faster than SPO.  Thin lines show matched replicates, thick lines their mean, and shaded regions one standard error.  Curves use a five-step trailing mean; Table~\ref{tab:main-results} reports unsmoothed aggregates.}
  \label{fig:training-curves}
\end{figure}

Our contributions are:
\begin{itemize}
  \item We identify a measure mismatch in trajectory whitening followed by a token-mean actor loss, and give action-token-measure normalization that preserves reward and credit granularity.
  \item We formulate a dispatch-causal, event-time prompt tracker whose estimate is invariant to receipt order for a fixed visible history.
  \item We evaluate \method{} across two ALFWorld model scales and Math-TIR, and isolate the normalization measure in a matched ablation.
\end{itemize}

\section{Method}
\label{sec:method}

\paragraph{Overview.}
\method{} keeps SPO's one-rollout dependency graph and terminal outcome unchanged.  A request reads and freezes a prompt baseline at dispatch; when it returns, its outcome is attached to the policy event that generated it.  The resulting scalar advantages are standardized under the action-token measure and then consumed by the same token-mean actor loss.  The two modifications therefore act at different boundaries: event-time memory governs which historical evidence defines the baseline, whereas measure alignment governs how a completed batch is standardized for optimization.

\subsection{Single-stream prompt values}

Let $x$ denote a repeatable task identity, $\tau_i$ a completed trajectory, and $R_i\in\{0,1\}$ its terminal outcome.  SPO maintains success and failure evidence $(\alpha_x,\beta_x)$ and reads a pre-update prompt value
\begin{equation}
  \widehat v_x=\frac{\alpha_x}{\alpha_x+\beta_x},
  \qquad A_i=R_i-\widehat v_x.
  \label{eq:spo-advantage}
\end{equation}
After consumption, evidence is updated with a policy-drift-dependent retention factor $\rho_i$:
\begin{equation}
  \alpha_x\leftarrow\rho_i\alpha_x+R_i,
  \qquad
  \beta_x\leftarrow\rho_i\beta_x+(1-R_i).
  \label{eq:spo-update}
\end{equation}
Each actor update uses one online rollout per prompt, while persistent evidence supplies its prompt baseline.  This single-stream dependency graph distinguishes SPO from group-based estimators.

\subsection{Event-time prompt memory}

Sequentially applying Equation~\ref{eq:spo-update} in completion order makes the tracker depend on systems timing.  A \emph{policy event} is the published actor snapshot used to generate a request.  Let $z_i$ denote the coordinate of the snapshot that generated trajectory $i$; successive snapshots advance this coordinate by the configured decay interval.  The pair $(\alpha_0,\beta_0)$ is the frozen offline evidence initialized at coordinate $z_0$.  \method{} freezes $\widehat v_x$ when a request is dispatched.  For a future dispatch $q$, let $\mathcal H_q(x)$ contain exactly the outcomes for task $x$ visible before that dispatch.  We form
\begin{equation}
  \begin{aligned}
  \alpha_q &= e^{-(z_q-z_0)}\alpha_0+
  \sum_{i\in\mathcal H_q(x)}e^{-(z_q-z_i)}R_i,\\
  \beta_q &= e^{-(z_q-z_0)}\beta_0+
  \sum_{i\in\mathcal H_q(x)}e^{-(z_q-z_i)}(1-R_i),
  \qquad
  \widehat v_q=\frac{\alpha_q}{\alpha_q+\beta_q}.
  \end{aligned}
  \label{eq:event-tracker}
\end{equation}
\begin{remark}[Receipt-order invariance]
For fixed $q$, $\mathcal H_q(x)$, and event coordinates, Equation~\ref{eq:event-tracker} is invariant to receipt permutation up to floating-point reduction error: reordering receipts only permutes two finite sums.  Receipt-order invariance is therefore conditional on the fixed visible history $\mathcal H_q(x)$.
\end{remark}

A returned outcome updates the baseline for subsequent dispatches; its own baseline remains the snapshot read at dispatch.

\paragraph{Implementation regime.}
The experiments dispatch a new request for a prompt after its previous request returns, while retaining first-finished completion across different prompts.  For each generation-policy event, the implementation advances $z$ by $-\log\rho$ with $\rho=0.875$; an outcome generated $k$ policy events before dispatch therefore receives weight $\rho^k$.  Equation~\ref{eq:event-tracker} is the event-indexed closed form of geometric retention, whereas the SPO baseline derives retention from post-update token drift.

\subsection{Action-token-measure normalization}

Let $n$ be the number of trajectories in a batch and $L_i$ the number of valid generated action tokens in trajectory $i$; observation and padding tokens have zero weight.  Trajectory-wise whitening defines
\begin{equation}
  \mu_{\mathrm{traj}}=\frac{1}{n}\sum_i A_i,
  \qquad
  \sigma^2_{\mathrm{traj}}=\frac{1}{n-1}\sum_i(A_i-\mu_{\mathrm{traj}})^2,
  \qquad
  \widetilde A_i^{\mathrm{traj}}=\frac{A_i-\mu_{\mathrm{traj}}}{\sigma_{\mathrm{traj}}+10^{-8}}.
  \label{eq:trajectory-normalization}
\end{equation}
It guarantees $\sum_i\widetilde A_i^{\mathrm{traj}}=0$, but a token-mean loss sees
\begin{equation}
  \frac{\sum_i L_i\widetilde A_i^{\mathrm{traj}}}{\sum_i L_i}
  =\frac{\operatorname{Cov}_n(L,\widetilde A^{\mathrm{traj}})}{\overline L},
  \label{eq:token-center}
\end{equation}
where $\overline L=n^{-1}\sum_iL_i$ and $\operatorname{Cov}_n$ uses divisor $n$.  The residual is therefore generally nonzero when token count and advantage co-vary.  \method{} instead standardizes under the same action-token measure consumed by the loss:
\begin{equation}
  \mu_{\mathrm{tok}}=\frac{\sum_iL_iA_i}{\sum_iL_i},
  \qquad
  \sigma^2_{\mathrm{tok}}=\frac{\sum_iL_i(A_i-\mu_{\mathrm{tok}})^2}{\sum_iL_i-1},
  \qquad
  \widetilde A_i^{\mathrm{tok}}=\frac{A_i-\mu_{\mathrm{tok}}}{\sigma_{\mathrm{tok}}+10^{-8}}.
  \label{eq:token-normalization}
\end{equation}
Every valid action token in one trajectory receives the same scalar $\widetilde A_i^{\mathrm{tok}}$.  The modification changes the normalization measure while preserving trajectory-level credit granularity.

\subsection{Optimization and asynchronous execution}

Both methods use the same uncertainty sampler, with prompt weight $\sqrt{\widehat v_x(1-\widehat v_x)}+0.05$, and collect the first completed requests within each policy version.  Old-policy work is drained when new parameters are published, and rollout importance ratios use an upper truncation threshold of two.  For negative advantage $A$, Dual-Clip caps the per-token surrogate loss at $-cA$; the evaluated SPO and \method{} recipes use $c=10$ and $c=2$, respectively.  Auxiliary reward, KL regularization, and moving-reference updates are disabled in both recipes.  Appendix~\ref{app:supplementary} reports the remaining implementation details.

\section{Experiments}
\label{sec:experiments}

\subsection{Protocol and metrics}

We compare SPO and \method{} in matched independent runs with identical model checkpoints, prompt sets, frozen offline prompt-value initializations, decoding, trajectory budgets, optimizer settings, and asynchronous runtime.  One plotted ALFWorld step averages 128 trajectories; one plotted Math-TIR step averages two 128-trajectory updates.  Thus 50 ALFWorld steps and 100 Math-TIR steps correspond to 6,400 and 25,600 online trajectories per method, respectively.  Our primary metric is normalized reward-curve area over the common budget: trapezoidal AUC for ALFWorld and the step mean (normalized rectangle-rule area) for Math-TIR.  The definitions differ only in endpoint weighting and are numerically near-identical on these curves; we retain each domain's original comparator.  Appendix~\ref{app:supplementary} gives both definitions.  The secondary metric is the mean reward over the final five plotted steps.  Differences are paired within matched replicates and always denote \method{} minus SPO; we report their mean and sample standard deviation.

\paragraph{Recipe variants.}
\method{} combines event-time memory and action-token-measure normalization with fixed retention $\rho=0.875$ and negative Dual-Clip $c=2$.  SPO uses drift-dependent retention and $c=10$.  Fixed retention gives policy events a learner-independent geometric coordinate.  The end-to-end comparison therefore measures the full recipes; only the normalization measure is isolated in Table~\ref{tab:norm-ablation}.  Dual-Clip is not ablated.

\paragraph{ALFWorld.}
We evaluate Qwen3.5-0.8B and Qwen3.5-2B agents \citep{qwen2026qwen35} on 128 canonical ALFWorld tasks \citep{shridhar2021alfworld}.  The agent receives the complete AgentLoop interaction history, and each model scale uses its own frozen prompt-value initialization.  Results are averaged over three matched runs at 0.8B and four at 2B; the task set, context, interaction limit, and online trajectory budget are otherwise identical.

\paragraph{Math-TIR.}
We evaluate Qwen3.5-0.8B with native Python tool use on a 1,500-example training split of DAPO-Math-17K \citep{yu2025dapo}.  Both methods share the same supervised cold-start checkpoint and prompt-value initialization.  Results are averaged over five matched runs.  Appendix~\ref{app:supplementary} gives the full protocol.

\subsection{Main results and training dynamics}

\begin{table}[h]
  \centering
  \caption{\method{} improves online learning efficiency in every evaluated setting.  Paired differences are reported in percentage points on the normalized $[0,1]$ scale, as mean $\pm$ sample s.d.; end reward averages the final five plotted steps.}
  \label{tab:main-results}
  \setlength{\tabcolsep}{9pt}
  \begin{tabular}{@{}lccc@{}}
    \toprule
    & \multicolumn{2}{c}{ALFWorld} & Math-TIR \\
    \cmidrule(lr){2-3}\cmidrule(l){4-4}
    & Qwen3.5-0.8B & Qwen3.5-2B & Qwen3.5-0.8B \\
    \midrule
    \spobase{} area & $0.532$ & $0.522$ & $0.217$ \\
    \method{} area & $\mathbf{0.722}$ & $\mathbf{0.681}$ & $\mathbf{0.242}$ \\
    \addlinespace[2pt]
    $\Delta$ area (points) & $+19.00\pm8.95$ & $+15.92\pm10.25$ & $+2.50\pm1.64$ \\
    $\Delta$ end reward (points) & $+7.86\pm7.18$ & $+4.88\pm6.83$ & $+5.03\pm4.74$ \\
    \bottomrule
  \end{tabular}
\end{table}

\paragraph{Overall performance and learning efficiency.}
Table~\ref{tab:main-results} and Figure~\ref{fig:training-curves} show a consistent advantage across model scales and task families.  Every ALFWorld run improves in curve area, and six of seven also finish with a positive end-reward difference.  Math-TIR shows a smaller positive mean gain, with both metrics improving in four of five runs.  The matched curves begin together and separate during training in all three settings.

\paragraph{When does measure mismatch matter?}
Equation~\ref{eq:token-center} predicts that trajectory whitening departs most from the actor's measure when response token count co-varies with advantage.  The larger gains on long-horizon ALFWorld and the smaller Math-TIR improvement are qualitatively consistent with this prediction.

\subsection{Ablation and optimization diagnostics}

\paragraph{Action-token-measure normalization.}
We isolate the normalization measure while holding the prompt tracker, retention, Dual-Clip, importance correction, optimizer, and the presence of standardization fixed.  Table~\ref{tab:norm-ablation} shows that action-token-measure normalization raises the short-horizon learning composite by $10.70\pm3.98$ percentage points relative to the otherwise matched trajectory-normalized variant.  Equation~\ref{eq:token-normalization} sets the actor-weighted mean $\sum_i L_i\widetilde A_i^{\mathrm{tok}}/\sum_i L_i$ to zero by construction.

\begin{figure}[h]
  \centering
  \includegraphics[width=\linewidth]{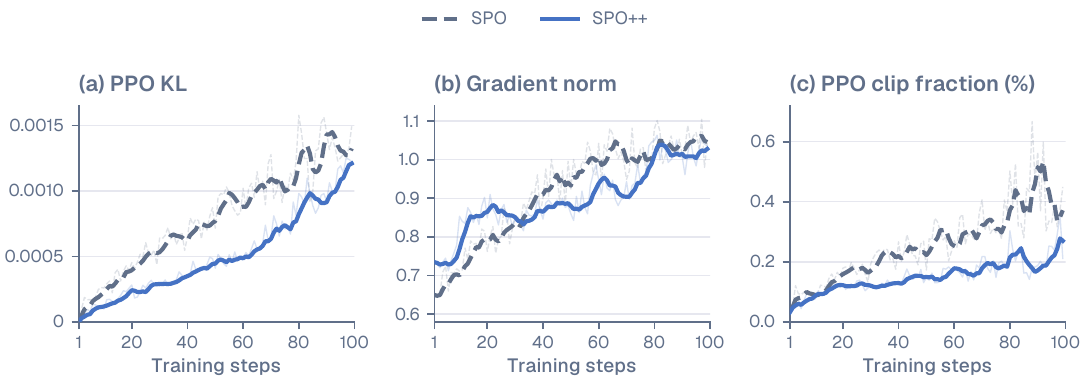}
  \caption{Recipe-level optimization diagnostics for SPO and \method{}.  Curves show one matched Math-TIR run with complete per-update logging; pale traces are raw updates and dark curves are five-step trailing means.  Other runs record a reduced diagnostic subset.}
  \label{fig:mechanism}
\end{figure}

\paragraph{Optimization diagnostics.}
Figure~\ref{fig:mechanism} compares the complete SPO and \method{} recipes.  \method{} exhibits lower PPO KL and clip fraction at a comparable gradient-norm scale.  Because retention and negative Dual-Clip also differ, these curves are recipe-level diagnostics rather than an isolated normalization effect.

\FloatBarrier

\begin{table}[h]
  \centering
  \caption{Matched normalization ablation on 0.8B ALFWorld over ten training steps and three seeds.  Values are changes in the learning composite $C_{10}$, in percentage points; the first two rows are relative to SPO and the last row is the paired difference.  We report mean $\pm$ sample s.d.}
  \label{tab:norm-ablation}
  \setlength{\tabcolsep}{7pt}
  \begin{tabular}{@{}lcc@{}}
    \toprule
    Variant & Normalization measure & $\Delta$ learning composite \\
    \midrule
    \method{} (trajectory norm.) & Trajectory & $+2.29\pm9.66$ \\
    \method{} & Action token & $\mathbf{+12.99\pm8.21}$ \\
    \addlinespace[2pt]
    Action-token vs. trajectory & Paired difference & $\mathbf{+10.70\pm3.98}$ \\
    \bottomrule
  \end{tabular}
\end{table}

\FloatBarrier

\section{Related Work}
\label{sec:related}

\paragraph{Single-rollout advantage estimation.}
PPO introduced clipped surrogate policy optimization \citep{schulman2017ppo}.  GRPO and subsequent large-scale refinements replace the critic with within-prompt relative comparison \citep{shao2024deepseekmath,yu2025dapo}; SPO instead carries prompt-level evidence across visits \citep{xu2025spo}.  Other single-rollout estimators obtain a baseline by sharing information across different prompts \citep{gong2026basis}, while MSSR stabilizes multimodal single-rollout RL through entropy-based advantage shaping \citep{liu2025mssr}.  Our work preserves SPO's persistent empirical prompt baseline and studies two consistency properties: matching standardization to the actor-loss measure and separating policy event time from learner receipt time.

\paragraph{Learned and implicit values.}
Value-based approaches improve advantage estimation by changing the source of the baseline.  VC-PPO pretrains a critic and decouples actor and critic GAE targets \citep{yuan2025vcppo}, while VAPO adds length-adaptive GAE and long-reasoning stabilizers \citep{yue2025vapo}.  $V_0$ predicts state-zero success with a frozen generalist value model \citep{zhang2026vzero}, and $V_{0.5}$ combines this prior with sparse empirical rollouts and adaptive sampling \citep{zhang2026vhalf}.  SAO and SAPO likewise learn values for single-response optimization, using separate and shared actor--value architectures, respectively \citep{hou2026sao,liang2026sapo}; BPCO instead stabilizes a bounded critic with Monte Carlo targets and retains raw residual advantages \citep{qi2026bpco}.

\paragraph{Objectives and supervision granularity.}
Other methods alter the objective or the granularity of supervision.  MaxRL moves the terminal-binary objective toward a compute-indexed likelihood approximation \citep{tajwar2026maxrl}; VIMPO derives token-level value recurrences from policy--reference log-ratios \citep{kang2026vimpo}; and SRPO turns hindsight reflections into dense token-level distillation targets \citep{liu2026srpo}.  Token-level loss aggregation is also an explicit stability choice in large-scale group-relative training \citep{yu2025dapo}.  Dr. GRPO identifies response-level length bias in GRPO \citep{liu2025drgrpo}, while Balanced Aggregation analyzes sign--length coupling induced by group-relative aggregation rules \citep{zeng2026balanced}.  These works change how response or token terms are aggregated.  Our change holds the token-mean loss and terminal reward fixed, and instead matches the measure used to standardize SPO's persistent scalar advantages.  The two design axes compose; \method{} remains critic-free and preserves the expected-reward objective.

\paragraph{Asynchronous policy optimization.}
Asynchronous LLM RL systems decouple rollout and optimization but must manage policy lag \citep{fu2025areal}.  PNPO reuses stale rollouts through prefix-aware ratios \citep{zhang2026pnpo}, A-3PO uses staleness-aware anchors \citep{li2025a3po}, and recent analyses expose old-policy mismatch and adapt trust regions \citep{guan2026missing,yang2026sat}.  We therefore separate policy event time, learner receipt time, and behavior-policy correction.  We evaluate on text-based ALFWorld and SPO's rule-verified TIR setting \citep{shridhar2021alfworld,xu2025spo,yu2025dapo}.

\enlargethispage{\baselineskip}

\section{Limitations}
\label{sec:limitations}

Our experiments study online learning efficiency on small Qwen3.5 models under limited training budgets; future evaluations can extend to longer horizons and out-of-distribution tasks.  The end-to-end comparison changes event-time tracking, retention, and Dual-Clip jointly, while the shorter 0.8B ALFWorld ablation isolates normalization under standardized advantages.  That ablation changes both centering and batchwise scaling, and does not decompose their individual effects.  The one-request same-prompt concurrency cap restricts event-time experiments to cross-prompt completion order.  Persistent prompt values require repeatable task identities and offline initialization, first-completed collection can favor shorter trajectories, and outcome-level credit retains the sparse-reward cold-start challenge.

\section{Conclusion}
\label{sec:conclusion}

We introduced SPO++, which aligns single-stream advantage normalization with the action-token measure consumed by a token-mean actor loss and organizes persistent prompt evidence by policy event time.  Across two ALFWorld model scales and Math-TIR, SPO++ improves reward-curve area over SPO.  A paired ablation identifies action-token-measure normalization as the strongest isolated contributor.  Single-stream RL couples dependency reduction to persistent statistics that must align with both the policy clock and the measure optimized by the actor.

\begingroup
\bibliography{references}
\bibliographystyle{plainnat}
\endgroup

\clearpage
\appendix
\section{Additional experimental details}
\label{app:supplementary}

\subsection{Metric definitions}

For ALFWorld rewards $r_1,\ldots,r_T$, normalized trapezoidal area is
\begin{equation}
  \operatorname{AUC}_{\mathrm{ALF}}=
  \begin{cases}
  r_1, & T=1,\\
  \displaystyle\frac{1}{T-1}\sum_{t=1}^{T-1}\frac{r_t+r_{t+1}}{2}, & T>1.
  \end{cases}
\end{equation}
The Math-TIR comparator uses the step mean $T^{-1}\sum_t r_t$, which is the normalized rectangle-rule area.  A matched comparison truncates both methods to their common number of completed reward points before computing either area or the final-five-collection mean.  Main-text aggregates include replicates in which both methods reach the common training budget.

For the ten-step normalization ablation, the learning composite is
\begin{equation}
  C_{10}=\frac{1}{2}\left(
  \operatorname{AUC}_{\mathrm{ALF}}(r_{1:10})
  +\frac{1}{5}\sum_{t=6}^{10}r_t\right).
\end{equation}
Thus, $C_{10}$ equally weights normalized trapezoidal reward-curve area and the final-five reward mean, matching the frozen screening rule.  We report $C_{10}$ only for the ten-step normalization ablation; Table~\ref{tab:main-results} reports reward-curve area and final-five reward separately.

\subsection{Training and implementation details}

\begin{table}[h]
  \centering
  \caption{Training configuration for the reported experiments.}
  \label{tab:full-config}
  \small
  \begin{tabular}{>{\raggedright\arraybackslash}p{0.30\linewidth}>{\raggedright\arraybackslash}p{0.29\linewidth}>{\raggedright\arraybackslash}p{0.27\linewidth}}
    \toprule
    Item & ALFWorld & Math-TIR \\
    \midrule
    Model & Qwen3.5-0.8B / 2B & Qwen3.5-0.8B (shared SFT) \\
    Training prompts & 128 canonical tasks & 1,500 prompts \\
    Offline value rollouts & $1{,}024$ (8 / prompt) & $12{,}000$ (8 / prompt) \\
    Online trajectories / method & $6{,}400$ & $25{,}600$ \\
    Update / publication & 128 / every update & 128 / every two updates \\
    Train / rollout GPUs & 2 / 6 & 4 / 4 \\
    Context / horizon & full AgentLoop / 50 interactions & full Python-tool / 8 turns \\
    Decoding & \multicolumn{2}{c}{$T=1$, top-$p=1$, top-$k=-1$} \\
    PPO clip / epochs / LR / WD & \multicolumn{2}{c}{$[0.2,0.28]$ / $1$ / $10^{-6}$ / $0.1$} \\
    KL / EWMA / entropy & \multicolumn{2}{c}{$0$ / off / $0$} \\
    Reward shaping / prompt concurrency & \multicolumn{2}{c}{none / 1} \\
    Rollout IS & \multicolumn{2}{c}{token upper-truncate at 2} \\
    \bottomrule
  \end{tabular}
\end{table}

\FloatBarrier

Across domains, SPO updates evidence in learner receipt order, derives retention from policy drift, whitens one scalar per trajectory, and uses negative Dual-Clip $c=10$.  \method{} freezes values at dispatch, inserts returns by policy event time with $\rho=0.875$, whitens under the action-token measure, and uses $c=2$.  Table~\ref{tab:full-config} summarizes the shared controls.

The Math-TIR cold start precedes online RL.  We sample eight native Python-tool trajectories per prompt from a Qwen3.5-4B teacher ($T=0.6$, top-$p=0.95$, top-$k=20$) and retain verifier-positive, structurally valid traces.  The retained 362 prompts are split 326/36 (764/73 examples).  We train all parameters for two bfloat16 epochs with global batch 8, learning rate $10^{-5}$, weight decay $0.01$, cosine decay, and maximum length 40,960.  The SFT loss is applied to the final assistant message.

Offline prompt values are Jeffreys-smoothed estimates of eight binary verifier outcomes per prompt (total Beta mass eight).  Each pair shares its initialization: the model-specific base policy for ALFWorld and the shared cold-start policy for Math-TIR.  All eight outcomes enter the initialization.  Aggregates retain finite pairs that reach their common budget.

\FloatBarrier

\end{document}